\documentclass[preprint,review,12pt]{elsarticle}

\usepackage{graphicx}
\usepackage{amsmath}
\usepackage{amssymb}
\usepackage{setspace}
\usepackage{enumerate}
\usepackage{xcolor}
\usepackage{hyperref}
\definecolor{deepgreen}{rgb}{0,0.5,0}
\hypersetup{
    colorlinks=true,
    citecolor={deepgreen}
}
\usepackage{color}
\usepackage{array}
\usepackage[ruled,linesnumbered,vlined]{algorithm2e}
\usepackage{multirow}
\usepackage{microtype}

\newcolumntype{C}[1]{>{\centering\let\newline\\\arraybackslash\hspace{0pt}}m{#1}}
\newtheorem{defn}{\noindent $\mathbf{Definition}$}[section]

\newtheorem{prop}[defn]{$\mathbf{Proposition}$}
\newtheorem{thm}[defn]{$\mathbf{Theorem}$}

\newcommand{\p}{\partial}

\journal{Pattern Recognition}

\begin{document}

\begin{frontmatter}

\title{Tooth morphometry using quasi-conformal theory}

\author[label1]{Gary P. T. Choi}
\author[label2]{Hei Long Chan}
\author[label3]{Robin Yong}
\author[label3]{Sarbin Ranjitkar}
\author[label3]{Alan Brook}
\author[label3]{Grant Townsend}
\author[label4]{Ke Chen}
\author[label2]{Lok Ming Lui\corref{cor1}}

\address[label1]{John A. Paulson School of Engineering and Applied Sciences, Harvard University, USA}
\address[label2]{Department of Mathematics, The Chinese University of Hong Kong, Hong Kong}
\address[label3]{Adelaide Dental School, The University of Adelaide, Australia}
\address[label4]{Department of Mathematical Sciences, The University of Liverpool, United Kingdom}
\cortext[cor1]{Corresponding author.}
\ead{lmlui@math.cuhk.edu.hk}

\begin{abstract}
Shape analysis is important in anthropology, bioarchaeology and forensic science for interpreting useful information from human remains. In particular, teeth are morphologically stable and hence well-suited for shape analysis. In this work, we propose a framework for tooth morphometry using quasi-conformal theory. Landmark-matching Teichm\"uller maps are used for establishing a 1-1 correspondence between tooth surfaces with prescribed anatomical landmarks. Then, a quasi-conformal statistical shape analysis model based on the Teichm\"uller mapping results is proposed for building a tooth classification scheme. We deploy our framework on a dataset of human premolars to analyze the tooth shape variation among genders and ancestries. Experimental results show that our method achieves much higher classification accuracy with respect to both gender and ancestry when compared to the existing methods. Furthermore, our model reveals the underlying tooth shape difference between different genders and ancestries in terms of the local geometric distortion and curvatures. 
\end{abstract}


\end{frontmatter}


\section{Introduction}
In anthropology, bioarchaeology and forensic science, a major problem is to obtain useful information from human remains.
While it is possible to extract the DNA from the remains, the genetic information may be degraded during excavation or decomposition \cite{Kaestle02}. Also, the extraction process may creates irreversible damages to the samples \cite{Mooder05}. To avoid the above-mentioned issues, one possible alternative approach is to analyze the shape of the remains. Unlike tissues and skins, which decay significantly over time, teeth are morphologically stable and resistant to degradation. Hence, the shape analysis of teeth is important for interpreting information of gender, ancestry and other identifiable factors.

Traditional morphometric methods have been extensively used for the study of the human tooth variation in terms of tooth size \cite{Alvesalo09}, tooth weight \cite{Schwartz05} etc.. To have a better understanding of the tooth shape variation, it is more desirable to consider landmark-based geometric morphometrics, which compares teeth based on prescribed anatomical landmarks such as their cusps and pits. Earlier methods in landmark-based geometric morphometrics such as the Procrustes superimposition \cite{Gower75} and thin plate spline (TPS) transformation \cite{Bookstein89} have been applied for studying the dental variation of different populations \cite{Hanihara05,Hanihara08,Polychronis13,Yong18}. However, a well-known limitation of these mapping methods is that in general neither the entire tooth shapes nor the landmarks can be exactly matched. This inaccuracy may compromise the comparison between the geometry of different tooth shapes. In recent years, conformal and quasi-conformal mappings have been considered for the analysis of medical and biological shapes such as brain cortical surfaces \cite{Lui07,Choi15}, hippocampi \cite{Lui10,Chan16}, vestibular systems \cite{Wen15}, carotid arteries \cite{Choi17} and insect wings \cite{Jones13,Choi18c}. In particular, Teichm\"uller map, a special type of quasi-conformal maps, is advantageous in the sense that it allows for exact landmark matching and is associated with a constant conformal distortion, as well as a natural metric called the Teichm\"uller distance. The Teichm\"uller distances between shapes, together with the differences in curvature of the shapes, serve as a powerful tool for capturing and quantifying shape variation.

\begin{figure}[t!]
\centering
\includegraphics[width=\textwidth]{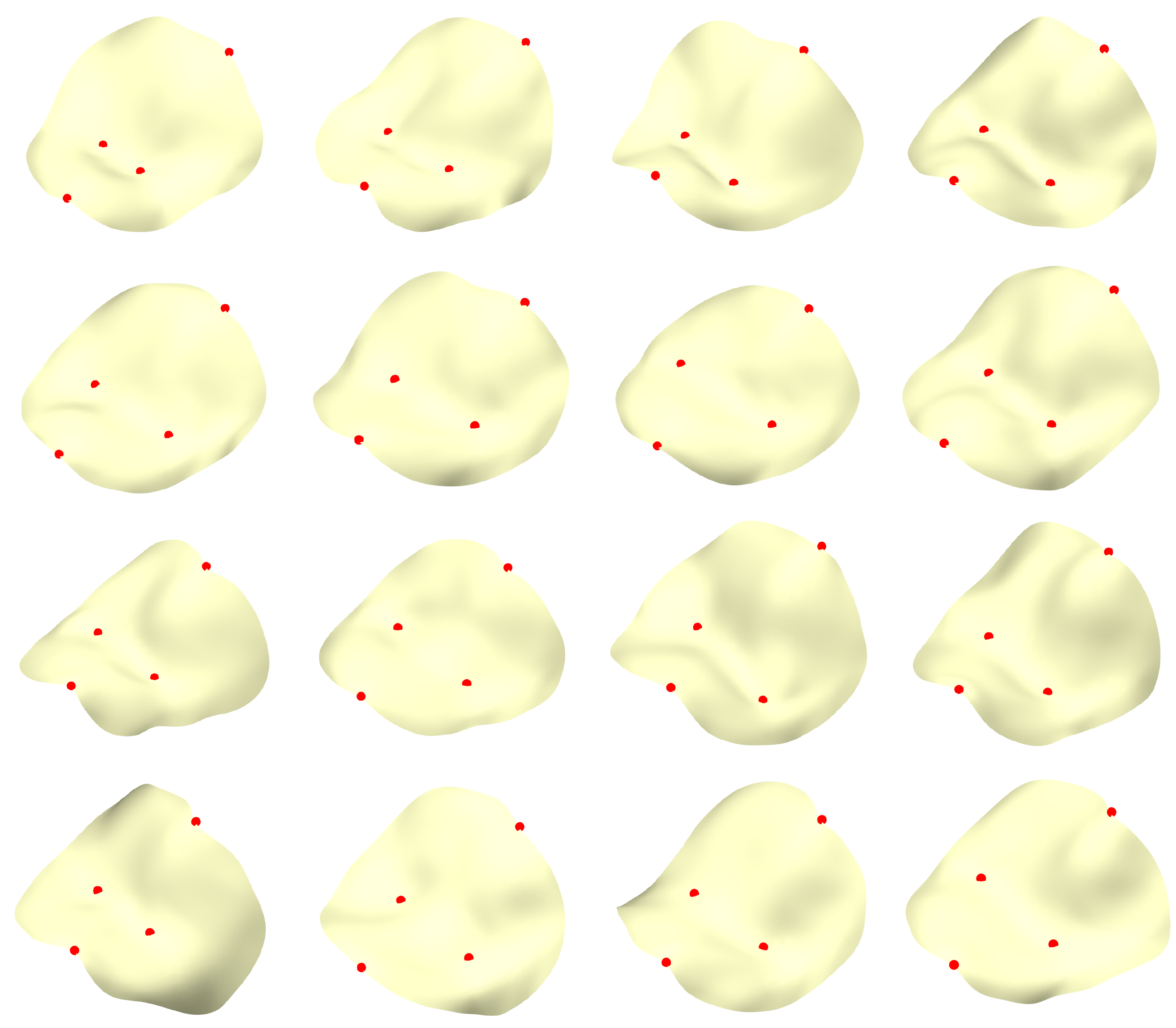}
\caption{Examples of the second upper premolar occlusal surfaces from two populations in Australia \cite{Brown11,Townsend15}, with four landmarks of the buccal cusp, the lingual cusp, the mesial fossa pit and the distal fossa pit highlighted in red. Each row shows four specimens with the same ancestry and gender. First row: Indigenous males. Second row: Indigenous females. Third row: European males. Fourth row: European females. It can be observed that the surfaces are different in terms of the overall shape, curvature and landmark positions.}
\label{fig:teeth}
\end{figure}

In this work, we propose a framework for accurately classifying a large set of 3D simply-connected open surfaces, by characterizing the shape variations using landmark-matching Teichm\"uller maps. The key to the unparalleled accuracy lies in taking into account the additional surface shape information using ideas from computational geometry and quasi-conformal theory. Illustration of our framework is done by applying the new algorithms to a dataset of tooth occlusal surfaces from Indigenous Australians \cite{Brown11} and Australians of European ancestry \cite{Townsend15} (see Figure \ref{fig:teeth} for examples). More specifically, to capture and quantify the shape differences between the 3D surfaces in terms of the overall shape, the curvature and the positions of the anatomical landmarks, we extend our previous work on landmark-matching Teichm\"uller map \cite{Meng16} to achieve an accurate 1-1 mapping between them, and further develop a quasi-conformal shape analysis model based on our previous work \cite{Chan16} for performing a classification. The classification results for the tooth dataset shed light on the ancestral variation and sexual dimorphism of teeth.

\section{Mathematical background}
We first review some important concepts in quasi-conformal theory. Readers are referred to \cite{Meng16,Lui14,Gardiner00} for more details.

\subsection{Quasi-conformal map}
Intuitively, quasi-conformal maps are orientation-preserving homeomorphisms with bounded conformality distortions. Under a quasi-conformal map, an infinitesimal circle is mapped to an infinitesimal ellipse with bounded eccentricity. The formal definition of quasi-conformal maps on the complex plane is given below.

\begin{defn}[Quasi-conformal maps]
A quasi-conformal map $f:\mathbb{C} \to \mathbb{C}$ is a map satisfying the Beltrami equation
\begin{equation}
\frac{\p f}{\p \bar{z}} = \mu_f(z) \frac{\p f}{\p z},
\end{equation}
for some complex-valued function $\mu_f(z)$ with $\|\mu_f\|_{\infty} < 1$.
\end{defn}
One can easily see that if $\mu_f = 0$, the above equation becomes the Cauchy-Riemann equation and hence $f$ is conformal (i.e. angle preserving).

More generally, let $S_1$, $S_2$ be two Riemann surfaces in $\mathbb{R}^3$. A Beltrami differential $\mu(z) \frac{\overline{dz}}{dz}$ on a Riemann surface $S$ is an assignment to each chart $(U_{\alpha}, \phi_{\alpha})$ on an $L_{\infty}$ complex-valued function $\mu_{\alpha}$, defined on local parameter $z_{\alpha}$ such that $\mu_{\alpha} \frac{\overline{dz_{\alpha}}}{dz_{\alpha}} = \mu_{\beta} \frac{\overline{dz_{\beta}}}{dz_{\beta}}$ on the domain which is also covered by another chart $(U_{\beta}, \phi_{\beta})$. An orientation-preserving diffeomorphism $f:S_1 \to S_2$ is said to be a quasi-conformal map associated with the Beltrami differential $\mu(z) \frac{\overline{dz}}{dz}$ if for any chart  $(U_{\alpha}, \phi_{\alpha})$ on $S_1$ and any chart $(U_{\beta}, \psi_{\beta})$ on $S_2$, the map $f_{\alpha \beta} = \psi_{\beta} \circ f \circ \phi_{\alpha}^{-1}$ is a quasi-conformal map.

In case the surfaces are simply-connected open surfaces, they can be represented by a single chart. Then, the computation of quasi-conformal maps between them can be easily reduced to the computation on the complex plane via a composition of mappings. Below is a useful property concerning the Beltrami coefficient associated with a composition of quasi-conformal maps, also known as the composition formula.
\begin{prop}[Composition of quasi-conformal maps]
If $f:\mathbb{C} \to \mathbb{C}$ and $g:\mathbb{C} \to \mathbb{C}$ are quasi-conformal maps, then $g \circ f$ is also a quasi-conformal map with Beltrami coefficient
\begin{equation}\label{eqt:composition}
\mu_{g \circ f}(z) = \frac{\mu_f(z) +\frac{\overline{f_z}}{f_z} \mu_g(f(z))}{1+\frac{\overline{f_z}}{f_z} \overline{\mu_f(z)}\mu_g(f(z))}.
\end{equation}
\end{prop}
From the above composition formula, it is easy to see that if $f$ is conformal and $g$ is quasi-conformal, then $\mu_{g \circ f}(z) = \mu_g(f(z))$ as $\mu_f = 0$. Also, if $f$ is quasi-conformal and $g$ is conformal, then $\mu_{g \circ f}(z) = \mu_f(z)$ as $\mu_g = 0$. In other words, the composition with a conformal map does not change the Beltrami coefficient.

\subsection{Teichm\"uller map}
Teichm\"uller map is a quasi-conformal map whose Beltrami coefficient has a constant norm. Hence, a Teichm\"uller map has a uniform conformal distortion over the entire domain. The formal definition of Teichm\"uller map is described below.

\begin{defn}[Teichm\"uller map]
Let $f:S_1 \to S_2$ be a quasi-conformal map. $f$ is said to be a \emph{Teichm\"uller map (T-map)} associated with the quadratic differential $q = \varphi dz^2$ where $\varphi:S_1 \to \mathbb{C}$ is a holomorphic function if its associated Beltrami coefficient is of the form
\begin{equation}\label{Teichmullermap}
\mu(f) = k \frac{\overline{\varphi}}{|\varphi|},
\end{equation}
for some constant $k < 1$ and quadratic differential $q \neq 0 $ with $||q||_1 = \int_{S_1} |\varphi| <\infty$.
\end{defn}

Furthermore, Teichm\"uller maps are closely related to a class of maps called extremal quasi-conformal maps.

\begin{defn}[Extremal quasi-conformal map]
Let $f:S_1 \to S_2$ be a quasi-conformal map. $f$ is said to be an \emph{extremal quasi-conformal map} if for any quasi-conformal map $h:S_1 \to S_2$ isotopic to $f$ relative to the boundary, we have
\begin{equation}\label{extremalmap}
K(f) \leq K(h),
\end{equation}
where $K(f)$ is the maximal quasi-conformal dilation of $f$. It is uniquely extremal if the inequality (\ref{extremalmap}) is strict when $h \neq f$.
\end{defn}

The two above-mentioned concepts are connected by the following theorem.

\begin{thm}[Landmark-matching Teichm\"uller map \cite{Reich02}]\label{landmarkteichmullerdisk}
\ Let $g:\partial \mathbb{D} \to \partial \mathbb{D}$ be an orientation-preserving diffeomorphism of $\partial \mathbb{D}$, where $\mathbb{D}$ is the unit disk. Suppose further that $g'(e^{i\theta}) \neq 0$ and $g''(e^{i\theta})$ is bounded. Let $\{l^k\}_{k=1}^n \in \mathbb{D}$ and $\{q^k\}_{k=1}^n \in \mathbb{D}$ be the corresponding interior landmark constraints. Then there exists a unique Teichm\"uller map $f:(\mathbb{D},\{l^k\}_{k=1}^n) \to (\mathbb{D}, \{q^k\}_{k=1}^n)$  matching the interior landmarks, which is the unique extremal extension of $g$ to $\mathbb{D}$. Here $(\mathbb{D},\{l^k\}_{k=1}^n)$ denotes the unit disk $\mathbb{D}$ with prescribed landmark points $\{l^k\}_{k=1}^n$.
\end{thm}

Therefore, besides equipped with uniform conformal distortion, Teichm\"uller maps are extremal in the sense that they minimize the maximal quasi-conformal dilation. Furthermore, Teichm\"uller maps induce a natural metric, called the \emph{Teichm\"uller distance} \cite{Gardiner00}, which can be used to measure the difference between two shapes in terms of local geometric distortion.

\begin{defn}[Teichm\"uller distance]
For every $i$, let $S_i$ be a Riemann surface with landmarks $\{p_i^k\}_{k=1}^n$. The \emph{Teichm\"uller distance} between $(f_i,S_i)$ and $(f_j,S_j)$ is defined as
\begin{equation}\label{eqt:t-metric}
 d_T((f_i, S_i), (f_j, S_j)) = \inf_{\varphi} \frac{1}{2} \log K(\varphi),
\end{equation}
where $\varphi:S_i \to S_j$ varies over all quasi-conformal maps with $\{p_i^k\}_{k=1}^n$ corresponds to $\{p_j^k\}_{k=1}^n$, which is homotopic to $f_j^{-1} \circ f_i$, and $K$ is the maximal quasi-conformal dilation.
\end{defn}

\section{Proposed method}
In this section, we describe our proposed method for accurately classifying a large set of 3D simply-connected open surfaces. To characterize the shape variation in terms of the surface geometry as well as the prescribed landmarks on them, we first propose a method for computing lanmdark-matching Teichm\"uller maps between 3D surfaces. Then, with the Teichm\"uller mapping results, we further propose a shape classification model based on quasi-conformal theory.

\subsection{Landmark-matching Teichm\"uller map between simply-connected open surfaces}
Denote two simply-connected open surfaces by $S_i$ and $S_j$, each with $n$ landmarks $\{l_i^1, \dots, l_i^n\}$ and $\{l_j^1, \dots, l_j^n\}$. We aim to quantify the difference between the two surfaces using a landmark-matching Teichm\"uller map $f_{ij}:S_i \to S_j$ that satisfies
\begin{equation}
f_{ij}(l_i^k) = l_j^k, k = 1, \dots, n.
\end{equation}
Unlike other methods such as radial basis function and spline-based methods, our approach takes both the overall shape and the landmarks of the surfaces into account, and is guaranteed by quasi-conformal theory.

\begin{figure}[t!]
\centering
\includegraphics[width=0.95\textwidth]{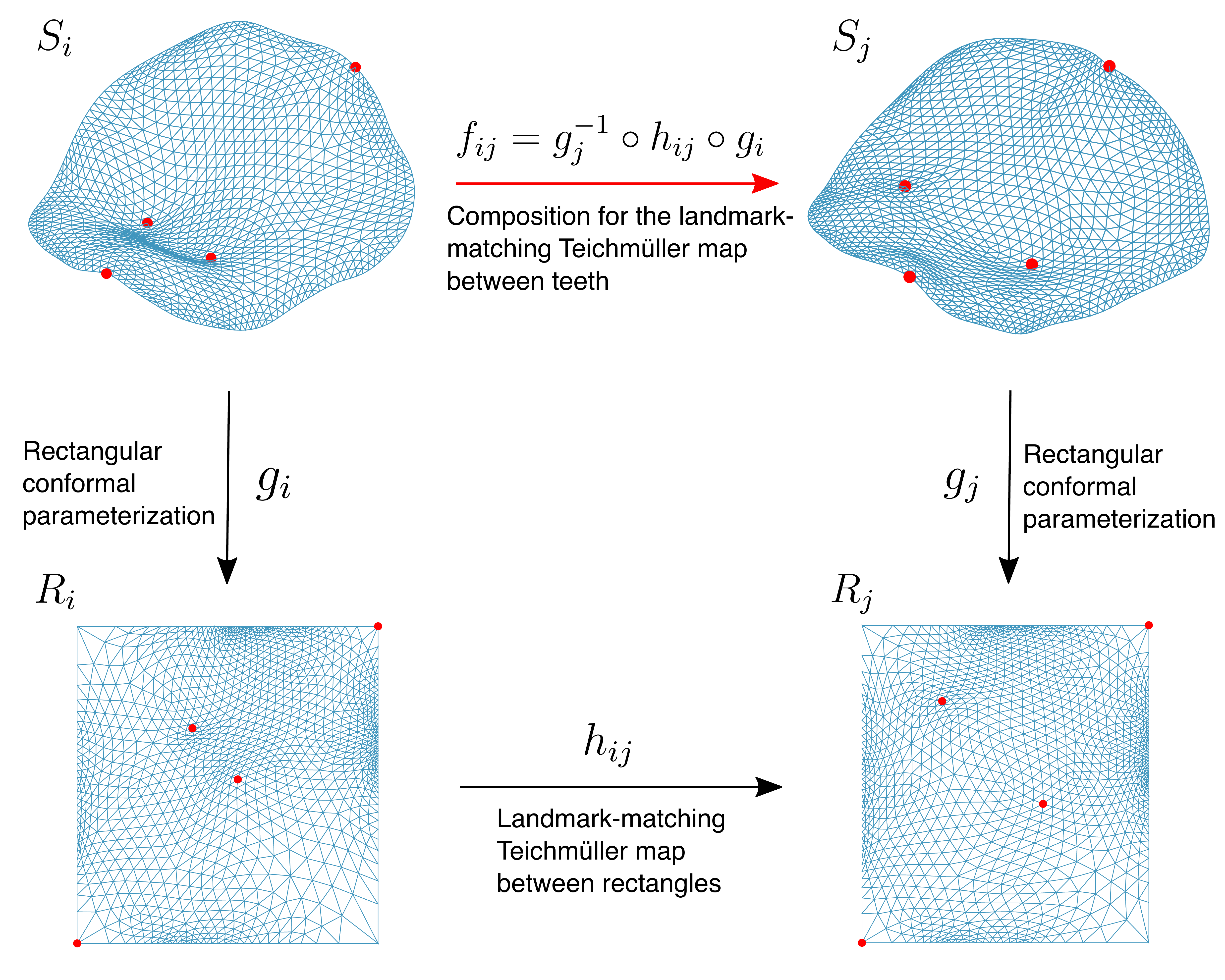}
\caption{An illustration of the computation of the landmark-matching Teichm\"uller map $f_{ij}$ between two occlusal surfaces  $S_i$ and $S_j$ (landmarks highlighted in red). The two surfaces are first flattened onto the plane by two rectangular conformal parameterizations $g_i$ and $g_j$. The landmark-matching Teichm\"uller map $h_{ij}$ between the two rectangles is then computed. Finally, the landmark-matching Teichm\"uller map $f_{ij}$ between the surfaces is given by the composition $g_j^{-1}\circ h_{ij} \circ g_i$.}
\label{fig:tmap_illustration}
\end{figure}

The procedure for finding $f_{ij}$ is outlined in Figure \ref{fig:tmap_illustration}. It consists of three steps, namely the rectangular conformal parameterizations, the landmark-matching Teichm\"uller map between the rectangles and the composition. Below, we discuss the technical detail of each step.

\subsubsection{Rectangular conformal parameterizations}
To simplify the mapping problem, we begin with flattening $S_i$ and $S_j$ onto the plane. While there exists other flattening methods such as area-preserving maps \cite{Choi18b,Choi18d}, conformal parameterizations are preferred in our case as they preserve the Beltrami coefficient and hence the conformal distortion under compositions. Following the approach in \cite{Meng16}, we compute two conformal maps $g_i:S_i \to {R_i}$ and $g_j:S_j \to {R_j}$ that flatten $S_i$ and $S_j$ onto two rectangular domains ${R_i}, {R_j}$ on the plane.

Note that the rectangular conformal parameterization algorithm in \cite{Meng16} was developed for point clouds. In our case of surface morphometry here, the approximation of the differential operators in \cite{Meng16} can be replaced by the mesh-based approximations, which are much simpler and more accurate. The rectangular conformal parameterization algorithm in \cite{Meng16} consists of a step of conformally parameterizing a surface onto the unit disk and a step of conformally mapping the unit disk to a rectnangle. Here, the disk conformal parameterization step can be replaced by our more recent disk conformal map algorithms \cite{Choi15b,Choi18a} for accelerating the computation and improving the accuracy.

\subsubsection{Landmark-matching Teichm\"uller map between the rectangular domains}
We then proceed to compute the landmark-matching Teichm\"uller map $h_{ij}: {R_i} \to {R_j}$ between the rectangular domains, following the approach in \cite{Meng16}. In particular, to satisfy the landmark correspondences, we require that
\begin{equation}
h_{ij}(g_i(l_i^k)) = g_j(q_i^k), k = 1, \dots, n.
\end{equation}
Again, note that \cite{Meng16} was developed for point clouds while the mesh structure is available in our case here. Therefore, the numerical algorithm used in \cite{Meng16} can be replaced by the more efficient mesh-based QC Iteration algorithm \cite{Lui14}.

Besides the landmark-matching Teichm\"uller map $h_{ij}$, we can also obtain the associated Beltrami coefficient $\mu_{h_{ij}}$. Since $h_{ij}$ is Teichm\"uller, $\mu_{h_{ij}}$ is with uniform norm, i.e. $|\mu_{h_{ij}}|$ is a constant over the entire domain.

\subsubsection{Composition for obtaining the landmark-matching Teichm\"uller map between the surfaces}
With the rectangular conformal maps $g_i, g_j$ and the landmark-matching Teichm\"uller map $h_{ij}$, a map $f_{ij}: S_i \to S_j$ can be obtained by
\begin{equation}
f_{ij} = g_j^{-1}\circ h_{ij} \circ g_i.
\end{equation}
Note that for any landmark $l_i^k$, we have
\begin{equation}
f_{ij}(l_i^k) = g_j^{-1}\circ h_{ij} \circ g_i (l_i^k) = g_j^{-1}(h_{ij}(g_i (l_i^k))) = g_j^{-1}(g_j(q_i^k)) = q_i^k.
\end{equation}
Hence, $f_{ij}$ is a landmark-matching map between $S_i$ and $S_j$.

Furthermore, the conformal distortion of $f_{ij}$ is the same as the conformal distortion of $h_{ij}$. In other words, $f_{ij}$ achieves a uniform conformal distortion $|\mu_{h_{ij}}|$ and hence $f_{ij}$ is a Teichm\"uller map. This can be explained by the composition formula \eqref{eqt:composition}. Since $g_i, g_j$ are conformal, we have $\mu_{g_i} = \mu_{g_j} = 0$. Now, by the composition formula, we have
\begin{equation}
\resizebox{.9 \textwidth}{!} 
{
$\mu_{h_{ij} \circ {g_i}}(z) = \frac{\mu_{g_i}(z) +\frac{\overline{{g_i}_z}}{{g_i}_z} \mu_{h_{ij}}({g_i}(z))}{1+\frac{\overline{{g_i}_z}}{{g_i}_z} \overline{\mu_{g_i}(z)}\mu_{h_{ij}}({g_i}(z))}= \frac{0 +\frac{\overline{{g_i}_z}}{{g_i}_z} \mu_{h_{ij}}({g_i}(z))}{1+0} = \frac{\overline{{g_i}_z}}{{g_i}_z} \mu_{h_{ij}}({g_i}(z)),$
}
\end{equation}
which implies that
\begin{equation}
|\mu_{h_{ij} \circ {g_i}}(z)| = \left|\frac{\overline{{g_i}_z}}{{g_i}_z} \mu_{h_{ij}}({g_i}(z)) \right| = |\mu_{h_{ij}} (g_i(z))| = |\mu_{h_{ij}}|.
\end{equation}
Similarly,
\begin{equation}
|\mu_{f_{ij}}(z)| = |\mu_{g_j^{-1} \circ h_{ij} \circ {g_i}}(z)| = |\mu_{h_{ij} \circ {g_i}}(z)| = |\mu_{h_{ij}} (g_i(z))| = |\mu_{h_{ij}}|.
\end{equation}

As a consequence, the Teichm\"uller distance is also uniquely determined by the maximal quasi-conformal dilation of the extremal map between the two rectangular domains. The Teichm\"uller distance $d$ between the two surfaces $S_i$ and $S_j$ is then given by
\begin{equation}
d_{ij} = \frac{1}{2}\log \frac{1+|\mu_{h_{ij}}|}{1-|\mu_{h_{ij}}|}.
\end{equation}
This completes the computation of the landmark-matching Teichm\"uller map between the two surfaces. The algorithm is summarized in Algorithm \ref{alg:tmap}.

\begin{algorithm}[h!]
\label{alg:tmap}
\KwIn{Two simply-connected open surfaces $S_i, S_j$ with landmarks $\{l_i^1, \dots, l_i^n\}$ and $\{l_j^1, \dots, l_j^n\}$.}
\KwOut{A landmark-matching Teichm\"uller map $f_{ij}: S_i \to S_j$, the Teichm\"uller distance $d_{ij}$.}
\BlankLine
Compute disk conformal parameterizations of $S_i$ and $S_j$ using the linear disk conformal map algorithm \cite{Choi18a}\;
Using the linear disk conformal map algorithm \cite{Choi18a} and the disk-to-rectangle conformal map algorithm \cite{Meng16}, obtain rectangular conformal parameterizations $g_i: S_i \to \mathbb{R}^2$ and $g_j: S_j \to \mathbb{R}^2$\;
Using the QC Iteration algorithm \cite{Lui14}, compute the landmark-matching Teichm\"uller map $h_{ij}: g_i(S_i) \to g_j(S_j)$ and obtain the Beltrami coefficient $\mu_{h_{ij}}$\;
Obtain $f_{ij} = g_j^{-1} \circ h_{ij} \circ g_i$ and $d_{ij} = \frac{1}{2}\log \frac{1+|\mu_{h_{ij}}|}{1-|\mu_{h_{ij}}|}$\;
\caption{Landmark-matching Teichm\"uller map between simply-connected open surfaces.}
\end{algorithm}

\begin{figure}[t]
\centering
\includegraphics[width=0.95\textwidth]{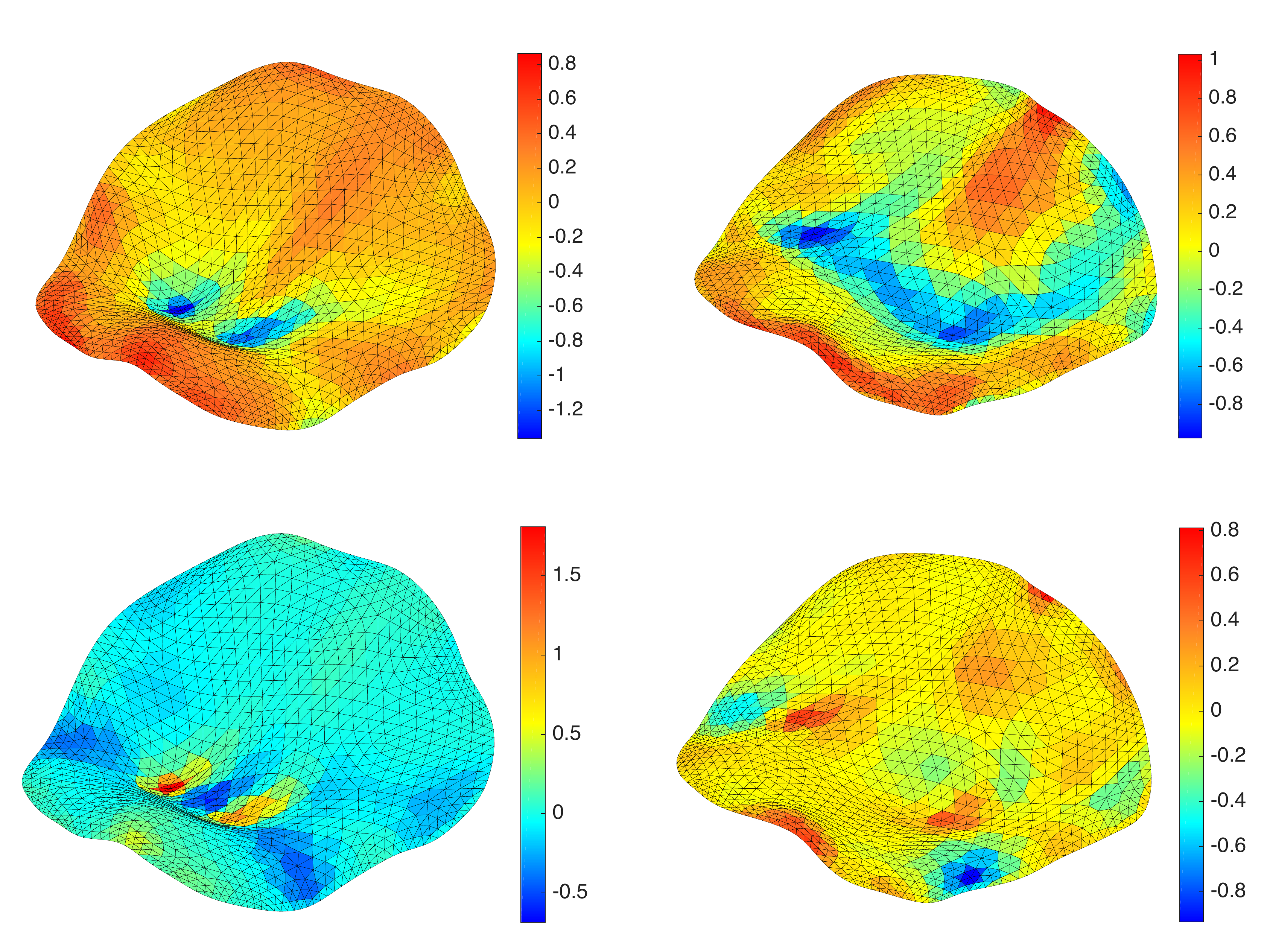}
\caption{Quantifying tooth geometry using mean and Gaussian curvatures. Top row: The mean curvature $H$ of two occlusal surfaces. Bottom row: The Gaussian curvature $K$ of them. An accurate comparison between the curvatures of different occlusal surfaces is made possible using landmark-matching Teichm\"uller maps.}
\label{fig:curvatures}
\end{figure}

\subsection{Quasi-conformal statistical shape analysis}
Note that the landmark-matching Teichm\"uller maps do not only provide us with a quantitative measure of the local geometric distortion of surfaces but also an accurate 1-1 correspondence between different parts of them. As illustrated in Figure \ref{fig:curvatures}, the mean and Gaussian curvatures also effectively quantify the surface geometry. With the aid of the landmark-matching Teichm\"uller maps, it is possible for us to analyze the surface shapes in terms of both the local geometric distortion and the curvature differences. Below, we devise a quasi-conformal statistical shape analysis model for building a surface classification machine.

Given a set of simply-connected open surfaces $\{S_i\}_{i=1}^N$, we first compute the landmark-matching Teichm\"uller maps $f_i: S_i \to S$ from every $S_i$ to their mean surface $S$. We can then obtain the associated Teichm\"uller distance $d_{i}$. Also, for each $S_i$, we compute the mean curvature $H_i$ and the Gaussian curvature $K_i$ at every vertex of it. After obtaining the results for all surfaces, a classification model can be built based on $d_i$, $H_i$, and $K_i$. More specifically, given a landmark-matching Teichm\"uller map $f_i: S_i \to S$, the following shape index $E_{\text{shape}}$ is considered:
\begin{equation}
E_{\text{shape}}(f_i)(v^k) = \alpha |H_i(v^k) - H(f_i(v^k))| + \beta |K_i(v^k) - K(f_i(v^k))| + \gamma d_i.
\end{equation}
Here $H$, $K$ represent the mean and Gaussian curvature of the mean surface $S$, $v^k$ are the vertices of $S_i$ with $k = 1, 2, \dots, M$, and $\alpha, \beta, \gamma$ are real nonnegative scalar parameters. Without loss of generality, we assume $\alpha^2+\beta^2+\gamma^2=1$. Note that $E_{\text{shape}}$ is a complete shape index for measuring all kind of distortion of the mapping $f_i$. The first two terms measure the curvature deviation of the mapping, and the third term measures the local geometric distortion of the mapping. In particular, $E_{\text{shape}} \equiv 0$ if and only if the two surfaces are identical up to rigid motion.

When compared to the formulation of shape index in \cite{Chan16}, the shape index $E_{\text{shape}}$ here consists of the same first two terms while the third term is different. More specifically, here we use the Teichm\"uller distance $d_i$ instead of the norm of the Beltrami coefficient $|\mu_i(v^k)|$ for the third term. Note that by quasi-conformal theory, $|\mu_i(v^k)|$ is always bounded by $[0, 1]$ for any bijective mappings. Instead, the Teichm\"uller distance is a metric and lies within $[0, \infty)$. As the first two terms $|H_i(v^k) - H(f_i(v^k))|$ and $|K_i(v^k) - K(f_i(v^k))|$ also have range $[0, \infty)$, using the Teichm\"uller distance as the third term gives a better balance between the three terms. Also, since $f_i$ is a Teichm\"uller map, $d_i$ is constant over the entire domain. Instead of the vertex-wise evaluation of $|\mu_i(v^k)|$, we can use a single scalar $d_i$ to capture the quasi-conformal distortion between $S_i$ and $S$.

Using the shape index function $E_{\text{shape}}$, a feature vector ${\bf c}_i = (c_{i}^1, c_{i}^2, ..., c_{i}^M)$ can be computed for each surface, with $c_{i}^k = E_{\text{shape}}(f_i)(v^k)$. Combining all feature vectors, we obtain a feature matrix
\begin{equation}
C = \begin{pmatrix}
{\bf c}_1 \\ {\bf c}_2 \\ \vdots \\ {\bf c}_N
\end{pmatrix}.
\end{equation}
The feature matrix provides full information of all shapes and hence can be used to develop a classification model. However, it is not necessarily true that all parts of the surfaces (i.e. all columns in $C$) are statistically significant for the classification. To extract the statistically significant regions that are the most related to the classification from the surfaces, the bagging predictors \cite{Leo96} are applied. We extract all vertices having a $p$-value less than or equal to a nonnegative threshold parameter $p_{cut} \in [0,1]$ as statistically significant regions. Readers are referred to \cite{Chan16} for more details.

Now, given a set of shapes and a binary classification criterion (e.g. classifying all tooth shapes into two ancestral/gender groups), we determine the optimal shape index parameters $(\alpha,\beta,\gamma)$ and the optimal threshold parameter $p_{cut}$ that yield the highest classification accuracy. To search for the optimal $(\alpha,\beta,\gamma)$, the following \emph{spherical marching scheme} (SMS) is utilized. Since we assume that $
\alpha^2+\beta^2+\gamma^2=1$, the space of the shape index parameters $\{(\alpha,\beta,\gamma)\in\mathbb{R}^3:\alpha^2+\beta^2+\gamma^2=1\}$ can be regarded as the unit sphere $\mathbb{S}^2$. Then, in order to search for the best set of parameters $(\alpha,\beta,\gamma)$ over $\mathbb{S}^2$ to maximize the classification accuracy in a timely manner, we parameterize $\mathbb{S}^2$ using the spherical coordinates
\begin{equation}
\resizebox{.9 \textwidth}{!} 
{
$
\mathbb{S}^2=\{(\text{sin}(\theta)\text{cos}(\varphi),\text{sin}(\theta)\text{sin}(\varphi),\text{cos}(\theta))\in\mathbb{R}^2:\theta\in[0,\pi],\varphi\in[0,2\pi)\}.
$
}
\end{equation}
Now, we discretize the parameter domain $[0,\pi]\times[0,2\pi)$ using regular gridding with density $\rho>0$, i.e.
\begin{equation}
\resizebox{.9 \textwidth}{!} 
{
$
[0,\pi]\times[0,2\pi)\approx \Omega =\left\{(n\rho,m\rho)\in\mathbb{R}^2:n=0,1,\dots,\frac{\pi}{\rho},m=0,1,\dots,\frac{2\pi}{\rho}\right\}.
$
}
\end{equation}
Then, for each $n,m$, $(n\rho,m\rho)$ corresponds to a set of parameters 
\begin{equation}
(\alpha,\beta,\gamma)_{n,m}=(\text{sin}(n\rho)\text{cos}(m\rho),\text{sin}(n\rho)\text{sin}(m\rho),\text{cos}(n\rho))
\end{equation}
on $\mathbb{S}^2$, and hence we can compute the classification accuracy of the proposed model using this set of parameters $(\alpha,\beta,\gamma)_{n,m}$. Therefore, the optimal $(\alpha,\beta,\gamma)$ can be chosen as the set of $(\alpha,\beta,\gamma)_{n,m}$ that gives the highest classification accuracy among all $n,m$. In practice, the density parameter $\rho$ is chosen within $[0.01\pi,0.03\pi]$. The optimal threshold parameter $p_{cut}$ for the extraction of statistically significant regions is determined by testing among different magnitudes of $10^k$, with $k = 0,-1,-2,-3,-4$. The quasi-conformal shape classification algorithm is summarized in Algorithm \ref{alg:classification}.

\begin{algorithm}[h!]
\label{alg:classification}
\KwIn{A set of simply-connected open surfaces $\{S_i\}_{i=1}^N$ with prescribed landmarks, and a classification criterion.}
\KwOut{The classification result and the optimal parameters $\alpha,\beta,\gamma,p_{cut}$.}
\BlankLine
Compute the mean surface $S$ of $\{S_i\}_{i=1}^N$\;
Compute the landmark-matching Teichm\"uller map $f_i:S_i \to S$ and the Teichm\"uller distance $d_i$ for all $i$\;
For all $i$ and for all $k$, evaluate the mean curvature difference $|H_i(v^k) - H(f_i(v^k))|$ and the Gaussian curvature difference $|K_i(v^k) - K(f_i(v^k))|$\;
Search for the optimal parameters $\alpha,\beta,\gamma,p_{cut}$ such that the shape index $E_{\text{shape}}$ and the statistically significant vertices together give the best classification result\;
\caption{Quasi-conformal shape classification.}
\end{algorithm}

It is noteworthy that the optimal shape index parameters $(\alpha,\beta,\gamma)$ determined by our model do not only maximize the classification accuracy with respect to a given criterion but also help us analyze the shape difference between the surfaces. More specifically, note that the mean and Gaussian curvatures uniquely determine a surface up to rigid motions, while the Teichm\"uller distance encodes the local geometric distortion. By changing the shape index parameters $(\alpha,\beta,\gamma)$ and comparing the corresponding classification accuracies, we can study the importance of each component (the mean curvature difference, the Gaussian curvature difference and the Teichm\"uller distance) for the classification and determine the major factor that distinguishes the surfaces.

\section{Data description}
\subsection{Study subjects}
Our study focuses on 140 subjects from two populations in Australia, namely the Indigenous group (subjects of Indigenous Australian ancestry) and the European group (subjects of European ancestry). The Indigenous group consists of 70 subjects (35 females, 35 males) of the Walpiri people (a group of Indigenous Australians who speak the Warlpiri language) living at Yuendumu in the Northern Territory of Australia \cite{Brown11}. The European group consists of 70 subjects (35 females, 35 males) with parents of Southern or Western European origin obtained from the Australian Twin study \cite{Townsend15}, with one co-twin from each twin pair selected randomly. The dental casts of the permanent dentitions of the subjects were obtained from the Yuendumu and Australian Twin collections housed in the Murray James Barrett Laboratory, Adelaide Dental School, The University of Adelaide. To overcome the problem of advanced tooth wear rate for Indigenous Australians due to hunter-gatherer dietary practices \cite{Molnar83} in the Yuendumu collection, assessment was limited to subjects in their early teens, with recently erupted premolars. Mean ages of the subjects were 12 years and 5 months (Indigenous females), 13 years (Indigenous males), 14 years and 8 months (European females), and 15 years and 7 months (European males). Readers are referred to \cite{Yong18} for a more detailed description of the dataset.

\subsection{Data acquisition and pre-processing}
The detailed procedure for the tooth data acquisition and the landmark protocol were described in \cite{Yong18}. The dental casts of the subjects were scanned using a 3D scanner at the resolution of 80-$\mu$m point distance. The upper second premolar in the maxillary right quadrant of each subject was extracted for this study. Four anatomical features on each tooth, including the buccal cusp, the lingual cusp, the mesial fossa pit and the distal fossa pit, were selected as landmarks by dentists. Besides the 4 landmarks, 88 curve and surface semi-landmarks were placed on each 3D tooth scan to delineate the occlusal circumference.

For our surface-based morphometric approach, it is desirable to represent the occlusal surfaces using triangle meshes. To achieve the triangle mesh representation, we first triangulated the landmarks and semi-landmarks of the occlusal surfaces. We then enhanced the mesh quality and resolution by surface remeshing \cite{Loop87}, thereby obtaining smooth, high-quality triangle meshes for our subsequent surface morphometry. Each remeshed occlusal surface consists of 1217 vertices. 

For each remeshed occlusal surface $S_i$, denote the four landmarks of the buccal cusp, lingual cusp, mesial fossa pit and distal fossa pit by $l_i^1, l_i^2,l_i^3, l_i^4$ respectively. Note that above-mentioned rectangular conformal parameterization procedure involves specifying four vertices on each occlusal surface to be mapped to the four corners of the corresponding rectangular domain. It is natural to consider the two crest landmarks $l_i^1, l_i^2$ on the boundary of the tooth surface as two corners, and the two other points on the boundary closest to the pit landmarks $l_i^3, l_i^4$ as the other two corners (see the bottom part of Figure \ref{fig:tmap_illustration} for an illustration). This ensures an accurate correspondence between the rectangular domains for different tooth surfaces.

\begin{figure}[t!]
\centering
\includegraphics[width=0.9\textwidth]{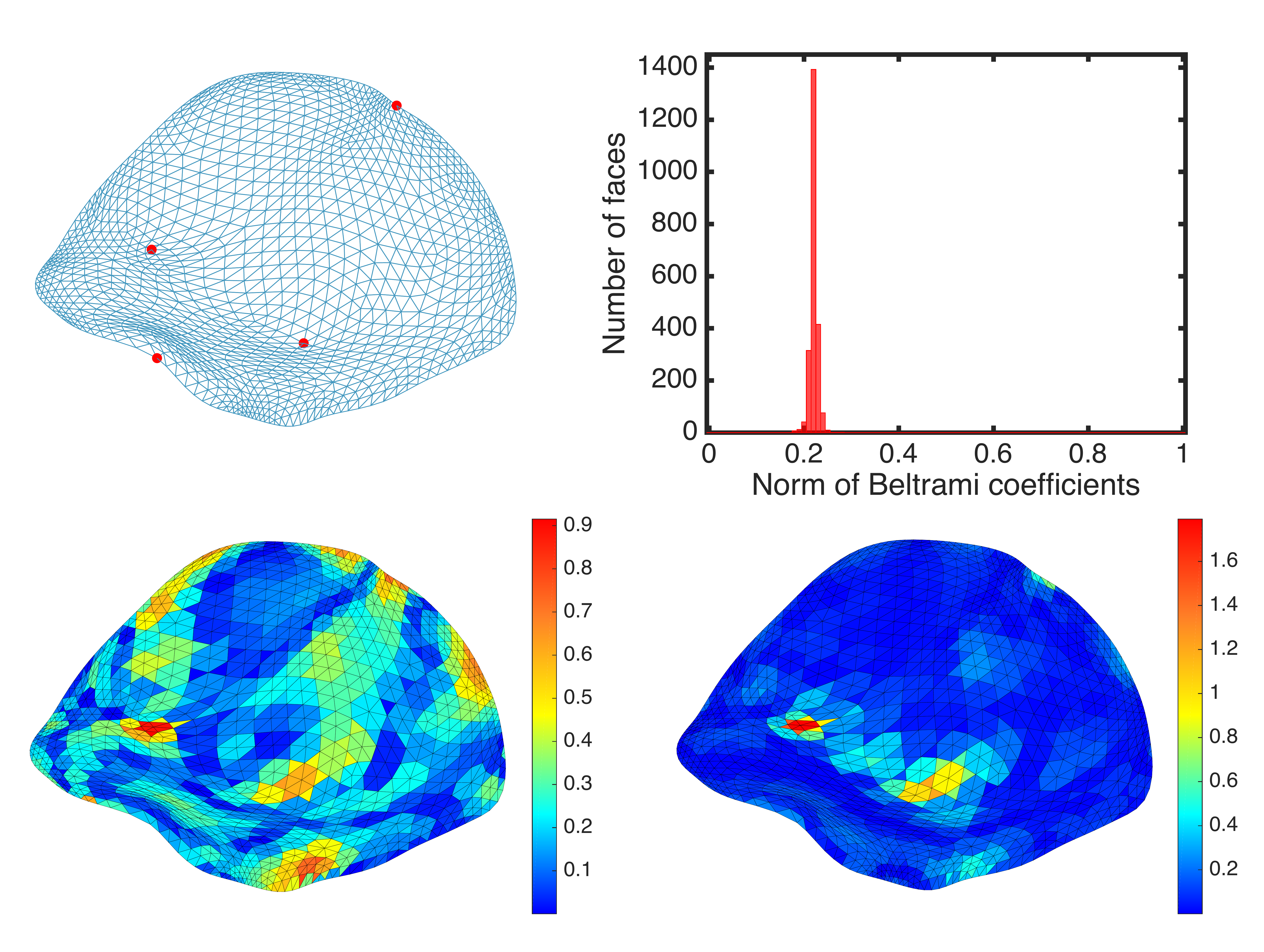}
\caption{The landmark-matching Teichm\"uller map between the two occlusal surfaces $S_i$, $S_j$ shown in Figure \ref{fig:tmap_illustration}. Top left: The landmark-matching Teichm\"uller mapping result $f_{ij}(S_i)$. Top right: The histogram of $|\mu_{f_{ij}}|$. Bottom left: The mean curvature difference $|H_i(v) - H_j(f_{ij}(v))|$ between the two occlusal surfaces. Bottom right: The Gaussian curvature difference $|K_i(v) - K_j(f_{ij}(v))|$ between the two occlusal surfaces.}
\label{fig:tmap_result}
\end{figure}

\section{Results}
\subsection{Landmark-matching Teichm\"uller map of occlusal surfaces}
As for a demonstration of our proposed method, we compute the landmark-matching Teichm\"uller map $f_{ij}$ between the occlusal surfaces $S_i$ and $S_j$ shown in Figure \ref{fig:tmap_illustration}. We remark that $S_i$ is an Indigenous male sample and $S_j$ is an European female sample. Figure \ref{fig:tmap_result} shows the mapping result and the curvature differences between the two surfaces. Comparing the mapping result in Figure \ref{fig:tmap_result} and the original surfaces shown in Figure \ref{fig:tmap_illustration}, it can be observed that $S_i$ is completely mapped onto $S_j$ under the mapping $f_{ij}$, with the landmarks exactly matched. The histogram of the norm of the Beltrami coefficients $|\mu_{f_{ij}}|$ is highly concentrated at one value, indicating that the mapping is Teichm\"uller. Also, using the landmark-matching Teichm\"uller map, we can easily evaluate the mean and Gaussian curvature differences between the two surfaces, thereby quantifying the shape difference between them. It is noteworthy that the major difference in Gaussian curvature is located at the fossa pits, while the difference in mean curvature is relatively widespread over the surfaces.

\subsection{Classification of the 140 upper second premolars with respect to ancestry and gender}
After demonstrating the effectiveness of the landmark-matching Teichm\"uller map for quantifying tooth shape difference, we deploy the mapping algorithm and the quasi-conformal statistical shape analysis model on the 140 upper second premolars in the dataset.

\subsubsection{The classification accuracy}
We first perform the classifications of all 140 occlusal surfaces in the dataset with respect to ancestry and gender using our proposed model. For comparison, we evaluate the classification accuracy achieved by our model as well as that achieved by two other classification methods respectively based on traditional morphometrics and landmark-based geometric morphometrics. More specifcally, we consider the area-based classification \cite{Colliot08,Chupin09} (note that the method in \cite{Colliot08,Chupin09} was originally volume-based for genus-0 surfaces, and so its analogue for simply-connected open surfaces is area-based) and the Procrustes-based classification \cite{Yong18}. 

Table \ref{table:classification} summarizes the classification results obtained by the two previous methods and our proposed method. It can be observed that the area-based method results in low classification accuracy for both classification tasks, which suggests that the traditional morphometric methods are incapable of capturing the tooth shape variation. The Procrustes-based method gives a satisfactory result for the classification with respect to ancestry but not gender. This implies that while earlier methods in landmark-based geometric morphometrics are more capable than the traditional morphometric methods, they are still insufficient for detecting certain kinds of tooth shape variation. In contrast to the two previous methods, our proposed method achieves $98.57\%$ accuracy (138 correct assignments out of 140 subjects) for the classification with respect to ancestry, and $97.14\%$ accuracy (136 correct assignments out of 140 subjects) for the classification with respect to gender. In both tasks, our method outperforms the existing methods. In particular, for the classification with respect to gender, the accuracy of our method is higher than the existing methods by around 30\%. This demonstrates the effectiveness of our proposed framework for tooth shape analysis.

\begin{table}[t!]
\small
\centering
\begin{tabular}{|C{25mm}||C{30mm}|C{30mm}|C{30mm}|} \hline
Classification Criterion & Overall Accuracy (Area-based \cite{Colliot08,Chupin09}) & Overall Accuracy (Procrustes-based \cite{Yong18})   & Overall Accuracy (Our Method)\\ \hline
Ancestry & 67.14\% & 91.43\%  & {\bf 98.57\%}\\ \hline
Gender & 51.43\%  & 68.57\%  & {\bf 97.14\%}\\ \hline
\end{tabular}
\caption{Classification accuracy for all the 140 upper second premolars with respect to ancestry and gender obtained by the area-based method \cite{Colliot08,Chupin09}, the Procrustes-based method \cite{Yong18} and our method.}
\label{table:classification}
\end{table}

\begin{table}[t!]
\scriptsize
\centering
\begin{tabular}{|C{17mm}|c|c|c|c||c|C{15mm}|C{14mm}|C{13mm}|} \hline
\multicolumn{5}{|c||}{Parameters} & \multicolumn{4}{c|}{Classification Result w.r.t. Ancestry} \\ \hline
Description & $\alpha$ & $\beta$ & $\gamma$ & $p_{cut}$ & $\#v$ & Correct Indigenous Rate & Correct European Rate & Overall Accuracy \\  \hline
Optimal & {\bf 0.1910} & {\bf 0.2034} & {\bf 0.9603} & {\bf 0.1} & {\bf 288} & {\bf 0.9857} & {\bf 0.9857} & {\bf 0.9857} \\  \hline
No $H$ term & 0 & 0.2034 & 0.9603 & \multirow{3}{*}{0.1} & 129 & 0.0286 & 0.9286 & 0.4786\\
No $K$ term & 0.1910 & 0 & 0.9603 &  & 108 & 0.6857 & 0.4571 & 0.5714\\
No $d$ term & 0.1910 & 0.2034 & 0 &  & 535 & 0.5429 & 0.8143 & 0.6786\\ \hline
\multirow{5}{*}{Varying $p_{cut}$} & 0.0922 & 0.9749 & 0.2028 & 0.0001 & 54 & 0.8286 & 0.8286 & 0.8286\\
& 0.2761 & 0.6974 & 0.6613 & 0.001 & 79 & 0.8429 & 0.8000 & 0.8214\\
& 0.1421 & 0.7449 & 0.6518 & 0.01 & 211 & 0.9714 & 0.9857 & 0.9786\\
&{\bf 0.1910} & {\bf 0.2034} & {\bf 0.9603} & {\bf 0.1} & {\bf 288} & {\bf 0.9857} & {\bf 0.9857} & {\bf 0.9857}\\
& 0.6956 & 0.1786 & 0.6959 & 1 & 1217 & 0.6571 & 0.7143 & 0.6857 \\  \hline
\end{tabular}
\caption{Classification results for all the 140 upper second premolars with respect to ancestry for various choices of the shape index parameters $\alpha$, $\beta$, $\gamma$ and the threshold parameter $p_{cut}$. Here, $\# v$ is the number of statistically significant vertices extracted by our model under the parameter settings. The correct Indigenous rate is calculated by $\frac{\text{\# of Indigenous subjects being classified as Indigenous}}{\text{Total \# of Indigenous subjects (i.e. 70)}}$, the correct European rate is calculated by $\frac{\text{\# of European subjects being classified as European}}{\text{Total \# of European subjects (i.e. 70)}}$, and the overall accuracy is evaluated over all the 140 subjects.}
\label{table:ancestry_parameters}
\end{table}

\subsubsection{The optimal parameters obtained by our model and their implications}
To have a better understanding, we analyze the optimal parameters obtained by our model for the two classification tasks. As shown in Table \ref{table:ancestry_parameters}, the optimal parameters for achieving the maximum classification accuracy with respect to ancestry are $(\alpha, \beta, \gamma) =  (0.1910, 0.2034, 0.9603)$, with $p_{cut} = 0.1$. From the values of $\alpha, \beta, \gamma$, it can be observed that the Teichm\"uller distance plays the most significant role in the classification with respect to ancestry. To study whether all the three terms (mean curvature difference, Gaussian curvature difference, Teichm\"uller distance) in the shape index are necessary for yielding an accurate classification, we consider setting one of $\alpha, \beta, \gamma$ to be 0 and evaluating the accuracy. We observe that dropping any of these terms will lead to a significant decrease in the accuracy. This implies that while the optimal $\alpha$ and $\beta$ are much smaller than $\gamma$, all the three terms are in fact important for the classification with respect to ancestry. In other words, the shape difference between the teeth from different ancestries is captured by the conformal (i.e. local geometric) distortion as well as the curvature differences.

Next, we consider varying the threshold parameter $p_{cut}$ and obtaining the best parameters $(\alpha, \beta, \gamma)$ that maximize the classification accuracy for different $p_{cut}$. In general, a larger $p_{cut}$ leads to a larger number of vertices identified as statistically significant by our model, and $p_{cut} = 1$ treats all vertices as statistically significant. Among several choices of $p_{cut}$, we observe that $p_{cut} = 0.1$ gives the highest classification accuracy. This suggests that using the entire surfaces does not necessarily lead to the best classification. Instead, it is important to extract certain regions on the surfaces which capture the shape difference between the Indigenous teeth and European teeth.

\begin{table}[t!]
\scriptsize
\centering
\begin{tabular}{|C{17mm}|c|c|c|c||c|C{15mm}|C{14mm}|C{13mm}|} \hline
\multicolumn{5}{|c||}{Parameters} & \multicolumn{4}{c|}{Classification Result w.r.t. Gender} \\ \hline
Description & $\alpha$ & $\beta$ & $\gamma$ & $p_{cut}$ & $\#v$ & Correct Male Rate & Correct Female Rate & Overall Accuracy \\  \hline
Optimal & {\bf 0.2330} & {\bf 0.0147} & {\bf 0.9724} & {\bf 0.001} & {\bf 468} & {\bf 0.9857} & {\bf 0.9857} & {\bf 0.9857} \\  \hline

No $H$ term & 0 & 0.0147 & 0.9724 & \multirow{3}{*}{0.01} & 1217 & 0.9429 & 0.9857 & 0.9643\\
No $K$ term & 0.2330 & 0 & 0.9724 &  & 478 & 0.9857 & 0.9857 & 0.9857\\
No $d$ term & 0.2330 & 0.0147 & 0 &  & 0 & N/A & N/A & N/A\\ \hline

\multirow{5}{*}{Varying $p_{cut}$} & 0.187 & 0.0118 & 0.9823 & 0.0001 & 185 & 0.9857 & 0.9857 & 0.9857\\
& {\bf 0.2330} & {\bf 0.0147} & {\bf 0.9724} & {\bf 0.001} & {\bf 468} & {\bf 0.9857} & {\bf 0.9857} & {\bf 0.9857}\\
& 0.0281 & 0.1093 & 0.9936 & 0.01 & 1188 & 0.9571 & 0.9857 & 0.9714\\
& 0.0351 & 0.1841 & 0.9823 & 0.1 & 1198 & 0.9429 & 0.9857 & 0.9643\\
& 0 & 0.9049 & 0.4258 & 1 & 1217 & 0.9714 & 0.9857 & 0.9786\\  \hline
\end{tabular}
\caption{Classification result for all the 140 upper second premolars with respect to gender for various choices of the shape index parameters $\alpha$, $\beta$, $\gamma$ and the threshold parameter $p_{cut}$. Refer to Table \ref{table:ancestry_parameters} for the description of the terms. }
\label{table:gender_parameters}
\end{table}

A similar analysis on the choices of the parameters can be performed for the classification with respect to gender (Table \ref{table:gender_parameters}). The optimal parameters for achieving the maximum accuracy are $(\alpha, \beta, \gamma) = (0.2330, 0.0147, 0.9724)$, with $p_{cut} = 0.001$. This time, it can be observed that the Teichm\"uller distance term is dominant in the shape index, while the Gaussian curvature difference term is with an extremely small weight. By setting one of $\alpha, \beta, \gamma$ to be zero, we can see that dropping the mean curvature difference term or the Gaussian curvature difference term in the shape index do not affect the classification accuracy much. By contrast, dropping the Teichm\"uller distance term will even lead to zero statistically significant vertices and hence the classification cannot be done. In other words, the shape difference between teeth from different genders is mostly captured by the local geometric distortion but not the curvature differences. Again, by varying $p_{cut}$ and evaluating the accuracy based on the corresponding optimal parameters, it can be observed that taking too many or too few vertices will lead to a sub-optimal result for the classification with respect to gender.

\subsubsection{The statistically significant regions on the occlusal surfaces for the two classification tasks}
\begin{figure}[t!]
\centering
\includegraphics[width=0.45\textwidth]{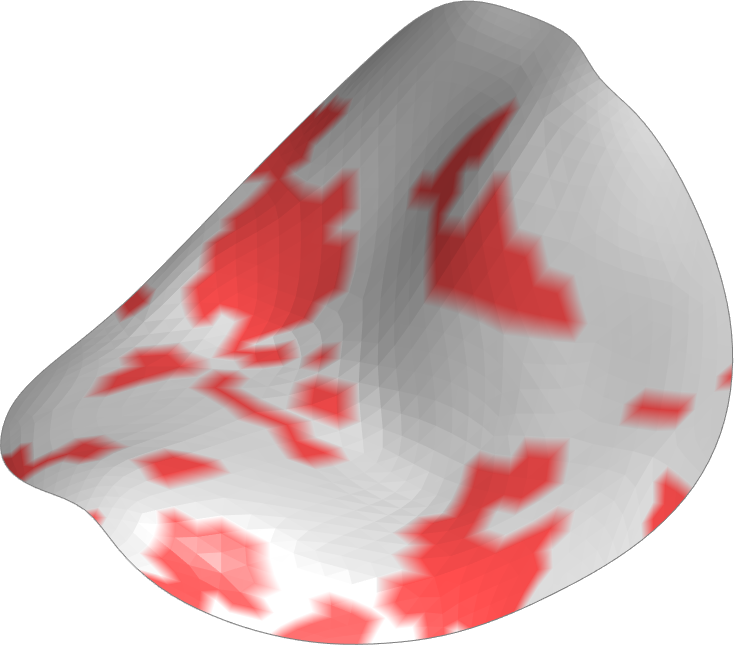}
\includegraphics[width=0.45\textwidth]{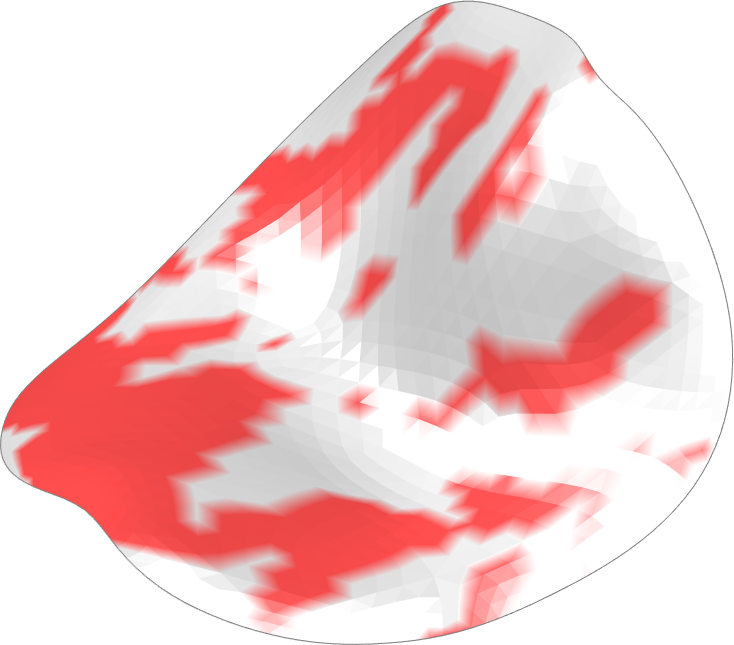}
\caption{The statistically significant regions (highlighted in red) extracted by our algorithm for the classifications with respect to ancestry (left) and gender (right), visualized on the mean surface of the 140 occlusal surfaces. }
\label{fig:significant}
\end{figure}

We compare the statistically significant regions identified by our proposed model for the two classification criteria. As recorded in Table \ref{table:ancestry_parameters} and Table \ref{table:gender_parameters}, around 20\% of the vertices (288 out of 1217 per surface) are statistically significant for the classification with respect to ancestry, while around 40\% (468 out of 1217 per surface) are statistically significant for the classification with respect to gender. In other words, the classification with respect to gender requires more global information. We visualize the regions by highlighting the relevent vertices in the mean surface of all teeth (see Figure \ref{fig:significant}). It can be observed that the statistically significant regions for the classification with respect to ancestry are primarily around the fossa pits, while those for the classification with respect to gender are primarily around the cusps.

\subsubsection{Possible explanation for the improvement achieved by our model when compared to the existing methods}
It is natural to ask why our method is capable of achieving a significant improvement in classification accuracy when compared to the existing methods, especially for the classification with respect to gender. In fact, this can possibly be explained by the optimal parameters obtained by our model for the two classification tasks.

Note that the Procrustes-based method \cite{Yong18} aligns the teeth by rigid motions and studies their shape difference. Since the mean and Gaussian curvatures uniquely determine a surface up to rigid motions, the shape information captured by the Procrustes approach can be considered as that captured by the two curvature terms in our shape index. As we have analyzed above, the Teichm\"uller distance is the only significant factor in the shape index for the classification with respect to gender. Therefore, with the consideration of the Teichm\"uller distance in our proposed model, it is reasonable that we can achieve a significant improvement in the classification accuracy with respect to gender. As for the classification with respect to ancestry, we have pointed out above that both the curvature differences and the the Teichm\"uller distance are important. Therefore, it is again reasonable that the Procrustes approach \cite{Yong18} achieves satisfactory accuracy, and our proposed model leads to an even better result.

\subsection{Classifications over subgroups}
Besides performing the classifications over the entire set of 140 subjects, we consider the classifications over subgroups. More specifically, we study whether the classification with respect to ancestry within each gender group and the classification with respect to gender within each ancestral group are similar to the ones over the entire set of 140 subjects.

\begin{table}[t!]
\small
\centering
\begin{tabular}{|C{30mm}||c|c|c|c||C{24mm}|} \hline
Gender Group (size = 70) & $\alpha$ & $\beta$ & $\gamma$ & $p_{cut}$ & Ancestry Classification Accuracy \\  \hline
Female & 0.1950 & 0.0661 & 0.9786 & 0.01 & 0.9714\\ \hline
Male & 0.0912 & 0.0234 & 0.9956 & 0.01 & 0.9714\\  \hline
\end{tabular}
\caption{The optimal parameters $\alpha, \beta, \gamma, p_{cut}$ and the accuracy of our proposed model for the classification with respect to ancestry within each gender group (each with size = 70).}
\label{table:ancestry_subgroup}
\end{table}

\begin{table}[t!]
\small
\centering
\begin{tabular}{|C{30mm}||c|c|c|c||C{24mm}|} \hline
Ancestral Group (size = 70) & $\alpha$ & $\beta$ & $\gamma$ & $p_{cut}$ & Gender Classification Accuracy \\  \hline
Indigenous & 0.0940 & 0.0829 & 0.9921 & 0.01 & 0.9714\\ \hline
European & 0.1702 & 0.1813 & 0.9686 & 0.01 & 0.9714
 \\  \hline
\end{tabular}
\caption{The optimal parameters $\alpha, \beta, \gamma, p_{cut}$ and the accuracy of our proposed model for the classification with respect to gender within each ancestral group (each with size = 70).}
\label{table:gender_subgroup}
\end{table}

We first consider the classification with respect to ancestry within each gender group (female/male, each with 70 subjects in total). For each gender group, we compute a landmark-matching Teichm\"uller map for each surface and repeat the classification procedure on the 70 mapping results for classifying the teeth with respect to ancestry. As shown in Table \ref{table:ancestry_subgroup}, our method achieves over $97\%$ classification accuracy for both gender groups. Also, in the two sets of optimal shape index parameters, $\gamma$ is much greater than $\alpha$ and $\beta$. This suggests that our findings for the classification with respect to ancestry over the entire dataset also hold when we consider the classification among females and males separately. In other words, the aforementioned shape difference between the two ancestries can be found in both genders.

We then consider the classification with respect to gender within each ancestral group (Indigenous/European, each with 70 subjects in total). As shown in Table \ref{table:gender_subgroup}, our method achieves over $97\%$ classification accuracy for both ancestral groups. Also, the optimal $\gamma$ are again much greater than $\alpha$ and $\beta$. This suggests that our findings for the classification with respect to gender over the entire dataset also hold when we consider the classification among the two ancestries separately. In other words, the aforementioned shape difference between the two genders can be found in both ancestries.

\section{Conclusion}
In this work, we have developed a framework for tooth morphometry using quasi-conformal theory. Landmark-matching Teichm\"uller maps are first used for finding a 1-1 correspondence and the Teichm\"uller distance between tooth surfaces. Then, a quasi-conformal statistical shape analysis model based on the Teichm\"uller distance and curvature differences is developed for building a classification scheme. We have deployed our method on a dataset of Australian upper second premolars. Our method achieves better classification accuracy with respect to both ancestry and gender when compared to the existing methods. Moreover, the optimal parameters and statistically significant regions obtained by our model for the classifications reveal the shape difference between teeth from different groups. For future work, we plan to perform a more comprehensive shape analysis on dentition using our proposed method, and further apply the framework for the study of other human organs.

\section*{Acknowledgment}
\noindent Ronald Lok Ming Lui was supported by HKRGC GRF (Project ID: 14303414). Gary P. T. Choi was supported by the Croucher Foundation.

\section*{Competing interests}
\noindent The authors have no competing interests to declare.\\

\vspace{1cm}

{\bf Gary P. T. Choi} is with the John A. Paulson School of Engineering and Applied Sciences, Harvard University. His research interests include computational geometry, mathematical modeling and medical imaging.

{\bf Hei Long Chan} is with the Department of Mathematics, The Chinese University of Hong Kong. His research interests include medical imaging, shape analysis and image segmentation.

{\bf Robin Yong} is with the Adelaide Dental School, The University of Adelaide. His research interests include dental anthropology and 3D imaging.

{\bf Sarbin Ranjitkar} is with the Adelaide Dental School, The University of Adelaide. His research interests include dental phenomics and craniofacial biology.

{\bf Alan Brook} is with the Adelaide Dental School, The University of Adelaide. His research interests include medical anthropology and biological anthropology.

{\bf Grant Townsend} is with the Adelaide Dental School, The University of Adelaide. His research interests include craniofacial biology and medical anthropology.

{\bf Ke Chen} is with the Department of Mathematical Sciences, The University of Liverpool. His research interests include mathematical imaging and numerical linear algebra.

{\bf Lok Ming Lui} is with the Department of Mathematics, The Chinese University of Hong Kong. His research interests include computational quasi-conformal geometry and medical imaging.

\end{document}